\documentclass[conference]{IEEEtran}
\usepackage{times}
\usepackage{graphicx}
\usepackage{subfigure} 
\usepackage{cite}
\usepackage{algorithm}
\usepackage{algorithmic}
\usepackage{multicol}
\usepackage{vwcol}
\usepackage{graphicx}
\usepackage{epstopdf}
\usepackage{verbatim}
\usepackage{color, soul}

\usepackage{amsmath, amsfonts, amssymb, amsthm}
\DeclareMathOperator*{\argmin}{arg\,min}
\DeclareMathOperator*{\argmax}{arg\,max}

\begin{document} 

\title{A Two-Stage Subspace Trust Region Approach\\ for Deep Neural Network Training}
 
\author{
 \IEEEauthorblockN{Viacheslav Dudar\IEEEauthorrefmark{1}, Giovanni Chierchia\IEEEauthorrefmark{2}, Emilie Chouzenoux\IEEEauthorrefmark{2}\IEEEauthorrefmark{3}, Jean-Christophe Pesquet\IEEEauthorrefmark{3}, Vladimir Semenov\IEEEauthorrefmark{1} }
 \IEEEauthorblockA{\IEEEauthorrefmark{1}Taras Shevchenko National University of Kyiv, Faculty of Computer Science and Cybernetics, Ukraine}
 \IEEEauthorblockA{\IEEEauthorrefmark{2}Universit\'e Paris Est, LIGM UMR 8049, CNRS, ENPC, ESIEE Paris, UPEM, F-93162 Noisy-le-Grand, France}
 \IEEEauthorblockA{\IEEEauthorrefmark{3}Center for Visual Computing, CentraleSupelec, University Paris-Saclay, Chatenay-Malabry, France}
}
 
 \maketitle

\begin{abstract} 
In this paper, we develop a novel second-order method for training feed-forward neural nets. 
At each iteration, we construct a quadratic approximation to the cost function in a low-dimensional subspace.
We minimize this approximation inside a trust region through a two-stage procedure:
first inside the embedded positive curvature subspace, followed by a gradient descent step. 
This approach leads to a fast objective function decay, prevents convergence to saddle points, and alleviates the need for manually tuning parameters. We show the good performance of the proposed algorithm on benchmark datasets.
\end{abstract} 


\section{Introduction}
Deep neural nets are among the most powerful tools in supervised learning, which have shown outstanding performances in various application areas \cite{Lecun2015}. Unfortunately, training such nets still remains a time-consuming task. It is thus of primary importance to design new optimization algorithms that allow one to perform this training in a more efficient manner.

Stochastic gradient descent (SGD) is one of the most popular algorithms for neural network training. However, being a first-order optimization scheme, SGD presents a number of pitfalls due to the nonconvex nature of the problem at hand. First, a proper learning rate can be difficult to select, causing SGD to slow down or even diverge if the stepsize is chosen too large. Additionally, the same learning rate applies to all weight updates, which may be suboptimal in a deep net, because of vanishing/exploding gradient problems. It is well known that in many cases the average norm of the gradient decays for earlier layers \cite{Bengio94}. Lastly, the algorithm can be trapped into one of the plateaus of low gradient length, which  slows down the learning process.

Numerous variants of SGD have been developed for circumventing the aforementioned issues \cite{Bottou16}. Several of them are grounded on well-known accelerated first-order schemes, such as momentum \cite{Qian1999} and Nesterov accelerated gradient \cite{Nesterov1983}, whereas others revolve around adaptive learning rate strategies, such as Adagrad \cite{Duchi2011}, Adadelta \cite{Zeiler2012}, RMSProp \cite{Tieleman2012}, and Adam \cite{Kingma14}, the latter being one of the fastest algorithms among first-order schemes. It was also shown that deep learning is possible with first-order methods in case of suitable initialization and proper schedule for momentum \cite{Sutskever2013}. 

Recently, a renewed attention has been paid to second-order optimization schemes, because of their ability to reach lower values of the error function compared with first order methods, in particular for deep autoencoders \cite{Hinton06} and recurrent neural nets. Martens \cite{Martens10} proposed a Hessian-free approach based on a conjugate gradient descent for minimizing a local second-order approximation of the error function limited to a data minibatch, resorting to damping for avoiding too large steps, coupled with a Levenberg-Marquardt style heuristics to update the damping parameter. The author successfully applied his method to deep autoencoder training, and recurrent nets training \cite{Martens11}. Vinyals and Povey \cite{Vinyals12} proposed to optimize the objective function within the Krylov subspace delineated by the previous iterate, the gradient, and products of powers of Hessian and gradient. Typically, the chosen dimensionality of the space ranges between $20$ and $80$. The resulting quadratic function in the $K$-dimensional space is minimized using $K$ iterations of BFGS. The authors reported significant speeds up compared with Hessian-free optimization. In contrast with the two previous methods, Dauphin \emph{et al.}\ \cite{Dauphin2014} proposed a saddle-free Newton approach that uses the exact Hessian matrix (instead of a Gauss-Newton approximation) within a Krylov subspace of moderate dimensionality. The authors show that the Hessian matrix in this subspace is usually not positive definite, but it suffices to replace the negative eigenvalues with their absolute values in order to make 
this Newton-like method saddle point repellent. The authors reported some improvements in the context of autoencoding training.

In this paper, we propose a neural network training algorithm that combines trust region \cite{More83,Yuan2015} with a subspace approach \cite{Chouzenoux15tsp, Erway2010}, thus satisfying the following requirements:
\begin{enumerate}
\item it exploits the second-order information in order to move in directions of low curvature;\\[-0.5em]
\item it uses as many learning rates as network layers, so as to update different blocks of weights at different speeds;\\[-0.5em]
\item it relies on an automatic procedure to optimally adjust the learning rates at each iteration.
\end{enumerate}

The paper is organized as follows. In Section~\ref{s:general}, we provide the general idea of our algorithm. In Section~\ref{s:subspace}, we discuss the subspace choice. In Section~\ref{s:trust_region}, we explain how to estimate the learning rates within the trust region. In Section~\ref{s:algorithm}, we put all these techniques together and detail the resulting algorithm. In Section~\ref{s:results}, we show the numerical results obtained with our approach. Finally, conclusions are drawn in Section~\ref{s:conclusion}.

\section{General Idea}\label{s:general}
Training a neural network amounts to finding a weight vector $w\in \mathbb{R}^N$ that minimizes a global error function $F\colon  \mathbb{R}^N \to  \mathbb{R}$, so yielding the optimization problem:
\begin{equation}
\operatorname*{minimize}_{w \in \mathbb{R}^N}\; F(w). 
\end{equation}
The error function $F$ is a sum of many nonconvex twice-differentiable terms, one for each input-output pair available in the training set. In a stochastic setting, the data are decomposed into minibatches. We denote by $F_j(w)$ the error function evaluated over the $j$-th minibatch, which can be viewed as a stochastic approximation of function $F$ as minibatches are randomly selected throughout the optimization process. In the following, we denote the batch (resp. minibatch) gradients by 
$$g(w)=\nabla F(w) \quad \mbox{(resp. $g_j(w)=\nabla F_j(w)$)},$$ 
and the batch (resp. minibatch) Hessians by
$$H(w)=\nabla^2 F(w) \quad \mbox{(resp. $H_j(w)=\nabla^2 F_j(w)$)}.$$

Consider the $K$-dimensional subspace $S$ spanned by some orthonormal vectors $d_0, \dots, d_{K-1}$ in $\mathbb{R}^N$, and let $V=\left[d_0\;\dots\;d_{K-1}\right] \in \mathbb{R}^{N\times K}$. Our proposal consists of updating the weight vector according to the following rule:
$$w\leftarrow w-\sum_{k=0}^{K-1}{\alpha_k d_k}=w-V\alpha,$$
where $\alpha={\left[\alpha_0,\dots,\alpha_{K-1}\right]}^T$ is a vector of learning rates.

A local quadratic Taylor expansion of the error function around the current point $w$ reads:
$$F_j(w+\Delta w)\approx F_j(w)+{g_j(w)}^T{\Delta (w)} +\frac{1}{2}{\Delta w}^T H_j(w) {\Delta w}.$$
By substituting $\Delta w=-V\alpha$, we get
$$F_j(w-V\alpha)-F_j(w)\approx-{r}^T\alpha +\frac{1}{2}{\alpha}^T B{\alpha}=Q(\alpha),$$
by introducing $B={V}^T H_j(w){V}$ and $r=V^T g_j(w)$.

Note that although $Q(\alpha)$ is a quadratic function of $\alpha$, curvature matrix $B$ is not necessarily positive definite when $F_j$ is nonconvex. 


The classical trust region method~\cite{Yuan2015} consists of minimizing a quadratic approximation to the cost function within
a ball around current point $w$, defined as 
$${\|\Delta w\|}^2\le{\epsilon}^2.$$ 
In the proposed subspace approach, the trust region corresponds to a Euclidean ball for the coefficients $\alpha$, defined as
$${\|\Delta w\|}^2={\|V\alpha\|}^2={\|\alpha\|}^2\le{\epsilon}^2.$$ 
Then, the main step of our approach is the minimization of the quadratic function inside the trust region, namely
$$\mathrm{find}\qquad \alpha^{*}=\argmin_{{\|\alpha\|}^2\le{\epsilon}^2}{Q(\alpha)}.$$
The trust region bound $\epsilon$ is then determined with backtracking and linesearch, with the aim to maximize the decay of $F_j$. 

Unfortunately, preliminary experiments suggested us that such a classical approach results in a relatively slow minimization process. 
In our view, a possible explanation for this fact is as follows. The location of the minimizer of $Q(\alpha)$ inside the trust region is mostly determined by the negative curvature directions of the quadratic form ${\alpha}^T B{\alpha}$ (in these directions $Q(\alpha)$ decreases most rapidly). Negative curvature directions are however not very reliable, because the objective function is usually bounded from below. In practice, this yields very small steps of the algorithm, as the trust region size is chosen so as to decrease the function (backtracking), and thus the resulting norm of the update is small. One more observation confirming this fact is that in the case when $B$ becomes positive definite, then the decrease of the function is of  higher order of magnitude compared with situations when a negative curvature is present. 

As suggested in \cite{Dauphin2014}, a possible solution could be to ignore all negative curvature directions. But in that case, the algorithm will hardly escape from saddle points. We experimented this strategy, and it already showed better results than the classical trust region approach, expecially for deep nets.

In this work, we propose instead a two-stage approach that combines the above strategies. At the first stage, we ignore negative curvature directions and address the trust region problem only for the subspace generated by positive curvature eigenvectors. With backtracking, we find a trust region size that allows us to decrease the function $F_j$, and we move to the point just found. At the new point, we re-compute the gradient and make gradient descent step. The stepsize is determined with linesearch and backtracking.


Thanks to the gradient descent step in the second stage, the proposed algorithm possesses the capability to move away from saddle points. Moreover, it should make fast progresses, because of large steps performed at the first stage in the subspace of positive curvature eigenvectors. 

\section{Choice of the Subspace}\label{s:subspace}
Although previous works \cite{Martens10,Dauphin2014} employ subspaces of relatively high dimensions (from 20 to 500), our experiments show that a much lower dimensionality is 
beneficial in terms of error decrease versus time. Indeed, each additional vector in the subspace requires the evaluation of a different Hessian-vector product at each iteration, which is obviously time consuming. 

Consider the minimalistic subspace generated by 2 vectors, namely the gradient and the previous iterate \cite{Chouzenoux15tsp}: 
\begin{equation}
S_2(w_n)=\textrm{span}\left\{g(w_n), w_n-w_{n-1}\right\}.\label{eq:sub}
\end{equation}
(For simplicity, the iteration index $n$ will be omitted in the following.)
For logistic regression problems, this subspace is enough to achieve relatively fast convergence, but for neural nets the situation is different. A possible explanation for this may be related to the vanishing/exploding 
gradient problem. Thus, we would desire to allow distinct learning rates for weights from different layers. 

Consider a feed-forward neural net with $L$ layers and some vector space basis vector $d_k\in S$ of a $K$-dimensional subspace $S$.  We can block-decompose $d_k$ and the weight vector $w$ as follows
$$d_k=\begin{bmatrix}d_k^0\\d_k^1\\ \vdots \\d_k^{L-1}\end{bmatrix}, \quad w=\begin{bmatrix}w^0\\w^1\\ \vdots \\w^{L-1}\end{bmatrix},$$
where blocks $d_k^l$ and $w^l$ correspond to layer $l$ of connections of the neural net. 
Then, the separation of learning rates $\alpha_k^l$ for each layer $l$ and vector $d_k$ results in the updates
$$w^l\leftarrow w^l-\sum_{k=0}^{K-1}\alpha_k^l d_k^l,$$
which is equivalent to consider a larger subspace generated by
$$\widetilde {d}_k^0=\begin{bmatrix}d_k^0 \\0\\ \vdots \\0\end{bmatrix}, \widetilde{d}_k^1=\begin{bmatrix}0\\d_k^1\\ \vdots \\0\end{bmatrix},\dots,\widetilde{d}_k^{L-1}=\begin{bmatrix}0\\0\\ \vdots \\d_k^{L-1}\end{bmatrix},$$
leading to
$$w\leftarrow w-\sum_{l=0}^{L-1}\sum_{k=0}^{K-1}\alpha_{k}^{l}\widetilde{d}_k^l.$$
The resulting update scheme is similar to SGD with momentum, but with separate learning rates for each layer and automatic choice of them at each iteration.

Note that our algorithm requires the vectors forming the subspace to be orthonormal. Since these vectors are nonzero only in one block, the task of orthonormalization is split into $L$ separate lower-dimensional subtasks. A similar argument applies for efficiently computing the Hessian-vector products. Indeed, the vectors to be multiplied are non-zero only in one block. We can thus avoid redundant calculations by carefully extending the popular $R$-technique \cite{Pearlmuter1994} to the sparse case.

\section{Minimization within the trust region}\label{s:trust_region}
The problem of finding the minimizer of a quadratic function inside an Euclidean ball has been well investigated. Here, we modify the classical algorithm \cite{More83} in order to apply our two-stage approach. Let us recall that the quadratic function is expressed as
$$Q(\alpha)=-{r}^T\alpha +\frac{1}{2}{\alpha}^T B{\alpha}, \quad \alpha\in \mathbb{R}^{2L},$$
with $K=2$ as in \eqref{eq:sub}. Suppose that the eigenvalues of $B$ are $\lambda_1\le\dots\le\lambda_{2L}$, and the corresponding eigenvectors are denoted by $v_1,\dots,v_{2L}$. Assume that the first positive eigenvalue in the list is $\lambda_{i_0}$ (we assume that there exists at least one positive eigenvalue, otherwise the first stage is not performed at all).
Let us define the vector $\widetilde{r}$ with components $\widetilde{r_i}={r^T v_i}$, $i\in \{1,\ldots,2L\}$. When the trust region is given by $\|\alpha\|\le\varepsilon$, we need to find $\lambda\ge0$ such that the matrix $B+\lambda I$ is positive definite, and $\| \left(B+\lambda I\right)^{-1}r\|=\epsilon$. Then, the minimal value inside the trust region is reached at $\alpha=\left(B+\lambda I\right)^{-1}r$. 
Matrix $B$ can be represented as:
$$B=\sum_{i=1}^{2L}{\lambda_i v_i v_i^T}.$$
When we need to compute the minimum in the subspace spanned by its positive eigenvectors, we just restrict the sum to
$$B_{+}=\sum_{i=i_0}^{2L}{\lambda_i v_i v_i^T}.$$
Then, in order to find $\lambda$, we solve the nonlinear equation:
$$\phi_{+}(\lambda)=\sum_{i=i_0}^{2L}\frac{\widetilde{r}_i^2}{{\left(\lambda_i+\lambda\right)}^2}={\varepsilon^2}$$
subject to $\lambda>-\lambda_{i_0}$. Since $\phi_{+}(\lambda)$ is monotonically decreasing and convex, 
we resort to Newton method. It is important to initialize it at the point $\lambda^{(0)}>-\lambda_{i_0}$ such that $\phi_{+}(\lambda^{(0)})>\epsilon^2$. In this way, sequence $\lambda^{(n)}$ will be monotonically increasing and will not jump from the region of interest $\lambda>-\lambda_{i_0}$. 

One more point to pay attention to is that this algorithm (for positive part) is applicable in the case when the global minimizer of the quadratic function given by
$$\alpha^{*}=\sum_{i=i_0}^{2L}{\frac{\widetilde{r_i}}{\lambda_i}v_i}, \quad \|\alpha^*\|^2=\phi_{+}(0)$$
 is outside the trust region we consider. For our algorithm this is always the case (see next section for details), so we can initialize $\lambda^{(0)}=0$ for Newton iterations. 

When $\widetilde{r}_1=0$,  the initial problem is more difficult, as the solution is not guaranteed to be unique. Fortunately, Nesterov \emph{et al.} \cite{Nesterov2006} suggested a simple way to avoid the difficulties arising in this case. We need to choose any index $k_0$ such that $v_1^{k_0}\ne0$ (in fact we search for the index of the maximum absolute value of $v_1^k$), and make the assignment $r^{(k_0)}\leftarrow r^{(k_0)}+\varepsilon_0$. It can be proven that as $\varepsilon_0\rightarrow0$ the minimum point for this shifted problem converges to some minimum point of the initial one.

When the value $\lambda$ is found, the minimizer $\alpha^{*}$  (for the positive curvature directions) in the trust region is given by
$$\alpha^{*}=\sum_{i=i_0}^{2L}{\frac{\widetilde{r_i}}{\lambda_i+\lambda}v_i}$$

\section {Details of the main algorithm}\label{s:algorithm}
Algorithm~\ref{alg:main} describes the general procedure in more details. It starts by randomly selecting a minibatch and dividing it into $L$ roughly equal sub-minibatches. Minibatches and subminibatches contain equal number of training elements from each class. The gradient is computed using the whole minibatch, while for each Hessian-vector product the sub-minibatches are used. So doing, the calculation of all $2L$ Hessian-vector products requires only about twice more time than computing the gradient. Note that a similar idea was implemented in the Hessian-free optimizer \cite{Martens10}. Then, the algorithm performs the two-stage procedure explained earlier.

We found out that backtracking and linesearch greatly increase the convergence speed. In both stages, after the first guess of $\alpha$, we perform fictive updates of weights and compute the real value of the minibatch function at this new point. When there is no decrease of value, we start decreasing the trust region size by a factor $0.5$. This process of shrinking the trust region is finite, because the gradient of $F_j$ is calculated exactly, and it can be shown that all updates calculated with trust region algorithm have an obtuse angle with the gradient. After a decrease of the function is obtained, we continue reducing the trust region by a factor $0.7$ to maximize the function decay.

In the second stage, we use the previous gradient descent step length as the initial guess, and we start with the point obtained in the first stage. Moreover, after the first guess of step length where we got a decrease of the function, we also start increasing it by a factor $1.3$, and stop when a decrease smaller than at previous tested size is obtained.

\begin{algorithm}[tb]
  \caption{Two-Stage Subspace Trust Region}
  \label{alg:main}
  \begin{algorithmic}
  \STATE Randomly initialize $w_0$
  \STATE $w_1\leftarrow w_0-\epsilon_0 g_0(w_0); \Delta_1\leftarrow\epsilon_0\|g_0(w_0)\|$
  \FOR{$j=1,2,\dots$}
  \STATE Calculate gradient $g_j(w_j)$
  \FOR{$l=0,\dots,L-1$}
  \STATE $\{d_0^l, d_1^l\}\leftarrow$ orthonormalize $\left\{g_j^l(w_j), w_j^l-w_{j-1}^l\right\}$
  \STATE Calculate $H_j d_0^l$ and $H_j d_1^l$ with sub-minibatches
  \ENDFOR 
  \STATE $V_j\leftarrow\{d_0^0,d_0^1,\dots,d_0^{L-1},d_1^0,\dots,d_1^{L-1}\}$
  \STATE 
  \STATE \textsl{Find Hessian $B$ and gradient $r$ for subspace:}
  \FOR{$k_1,k_2=0,1$ and $l_1, l_2=0,\dots,L-1$}
  \STATE $B[k_1 L+l_1][k_2 L+l_2]\leftarrow\left(d_{k_1}^{l_1}, H_j d_{k_2}^{l_2}\right)$
  \STATE $r[kL+l]=\left(g_j(w_j), d_{k}^l\right)$
  \ENDFOR 
  \STATE Find $\{\lambda_1,...,\lambda_{2L}\}, \{v_1,\dots,v_{2L}\}$ for $B$
  \STATE
  \STATE \textsl{First stage (positive curvature step):}
  \IF {$\lambda_{2L}>0$}
  \STATE $\alpha^{*}\leftarrow\sum_{i=i_0}^{2L}{\frac{\widetilde{r_i}}{\lambda_i}v_i}$, $\Delta\leftarrow\|\alpha^*\|$
  \STATE Define operator $\alpha^{*}(\Delta)\leftarrow\argmin_{{\|\alpha\|}\le \Delta}{Q_{+}(\alpha)}$ 
  \IF {$F_j(w_j-V_j\alpha^{*})>F_j(w_j)$}
  \STATE \textbf{Do} $\Delta\leftarrow0.5\Delta$ \textbf{until} $F_j(w_j-V_j\alpha^{*}(\Delta))<F_j(w_j)$
  \ENDIF
  \STATE \mbox{\textbf{Do} $\Delta\leftarrow0.7\Delta$ \textbf{until}} \\ \mbox{$F_j(w_j-V_j\alpha^{*}(0.7\Delta))>F_j(w_j-V_j\alpha^{*}(\Delta))$}
  \STATE $w_j\leftarrow w_j-V_j\alpha^{*}(\Delta)$
  \ENDIF
  \STATE 
  \STATE \textsl{Second stage (gradient descent step):}
  \IF {$\lambda_{2L}>0$}
  \STATE Recalculate $g_j(w_j)$
  \ENDIF
  \IF {$F_j(w_j-\Delta_1\frac{g_j}{\|g_j\|})>F_j(w_j)$}
  \STATE \textbf{Do} $\Delta_1\leftarrow0.5\Delta_1$ \textbf{until} $F_j(w_j-\Delta_1\frac{g_j}{\|g_j\|})<F_j(w_j)$
  \STATE \mbox{\textbf{Do} $\Delta_1\leftarrow0.7\Delta_1$ \textbf{until}} \mbox{$F_j(w_j-0.7\Delta_1\frac{g_j}{\|g_j\|})>F_j(w_j-\Delta_1\frac{g_j}{\|g_j\|})$}
  \ELSE
  \STATE  \mbox{\textbf{Do} $\Delta_1\leftarrow1.3\Delta$ \textbf{until}}\\ \mbox{$F_j(w_j-1.3\Delta_1\frac{g_j}{\|g_j\|})>F_j(w_j-\Delta_1\frac{g_j}{\|g_j\|})$}
  \ENDIF
  \STATE $w_j\leftarrow w_j-\Delta_1\frac{g_j}{\|g_j\|}$
  
   \ENDFOR 
  \end{algorithmic}
\end{algorithm}
  
\section{Experimental results}\label{s:results}
In order to assess the performance of our algorithm, we considered the training of fully-connected multilayer neural networks with MNIST, a benchmark dataset of handwritten digits. First, we performed some experiments to show the validity of the proposed two-stage trust region procedure. Secondly, we compared our approach with two popular methods: Adam \cite{Kingma14} and RMSProp \cite{Tieleman2012}.

\subsection{Two-stage trust region assessment}
In our first experiments, we investigate how to handle negative eigenvalues of the matrix $B$, by testing the following approaches:

\textbf{Trust region} - A minimizer of $Q(\alpha)$ is found subject to $\|\alpha\|\le\varepsilon$. Backtracking and linesearch are used to determine the optimal value of $\varepsilon$.

\textbf{Only positive} - The coefficients of $\alpha$ are chosen from the subspace generated by the eigenvectors of $B$ corresponding to positive eigenvalues. Negative eigenvalues are ignored. 

\textbf{Saddle free} - Negative eigenvalues of $B$ are replaced with their absolute values, then the trust region method is applied.

\textbf{Positive-negative} - A minimum inside the trust region for the positive eigenvector subspace is found. After that we move to this new point, recompute the gradient, and consider the subspace generated by negative eigenvectors. Trust region sizes at both stages are determined with linesearch.

\textbf{Negative-positive} - Same procedure as Positive-Negative, except that the order of first and second stages is inverted.

\textbf{Two-stage} - The proposed approach. A step in the positive subspace is followed by a gradient descent step.

\begin{figure}[t]
\centering
\includegraphics[width=0.95\linewidth]{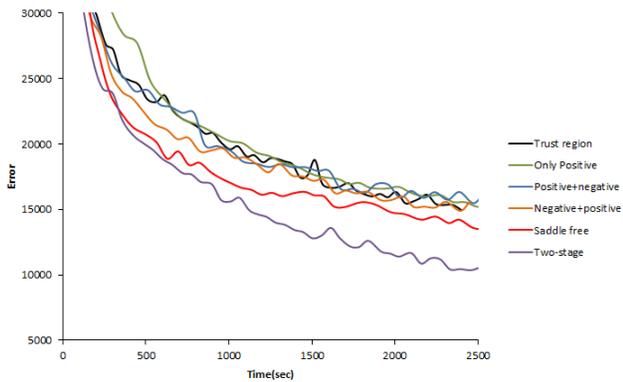}
\caption{Comparison of second-order methods for a 784-50-10 network.}
\label{fig:ex1}
\end{figure}

\begin{figure}[t]
\centering
\includegraphics[width=0.95\linewidth]{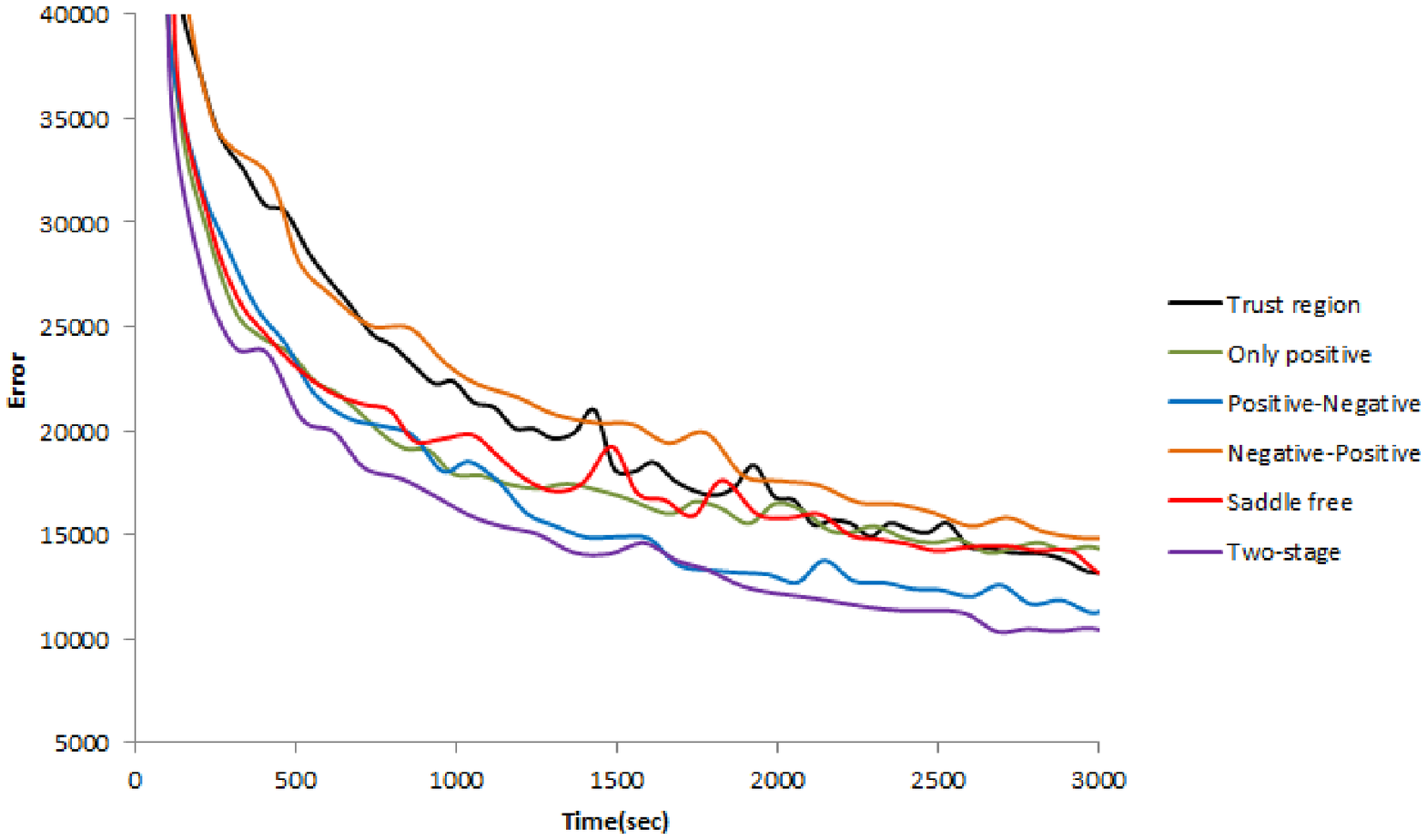}
\caption{Comparison of second-order methods for a 784-50-50-10 network.}
\label{fig:ex2}
\end{figure}

We tested the above approaches for a 2-layer net with 50 hidden units (784-50-10), and 3-layer net with 784-50-50-10 architecture. The same subspace and backtracking/linesearch algorithms were used for all tested methods. We used softmax output, tanh hidden functions, and the cross-entropy error function. This error function was measured after each epoch on the whole training set. We also used quadratic regularization with coefficient $10^{-4}$. The results are shown in Figs.~\ref{fig:ex1} and \ref{fig:ex2}, indicating that the proposed two-stage strategy exhibits the best performance.

%

\begin{figure}[h]
	\centering
	\includegraphics[width=0.95\linewidth]{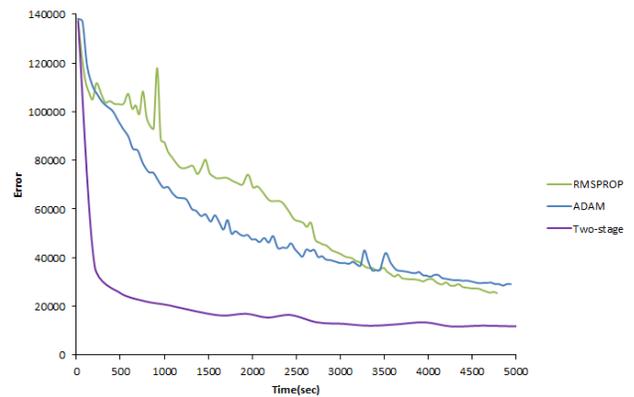}
	\caption{Comparison for a 784-80-70-60-50-40-30-20-10 network.}
	\label{fig:ex4}
\end{figure}

\subsection{State-of-the-art comparison}
We also compared the proposed approach with two popular first-order methods: Adam \cite{Kingma14} and RMSProp \cite{Tieleman2012} on a 8-layer net with 784-80-70-60-50-40-30-20-10. We again used tanh units, softmax output, and the cross-entropy error function. We also used sparse initialization, which  prevents saturation of learning at the beginning \cite{Sutskever2013}. Fig.~\ref{fig:ex4} reports the value of the error function versus the elapsed time. This experiment shows that our approach performs much better than first order methods on such deep network. This happens because 
the second-order information exploited by our algorithm, despite requiring more computations per iteration w.r.t.\ first-order methods, pays off with larger updates on deep networks.

\section{Conclusion}\label{s:conclusion}
In this work, we have proposed a two-stage trust region subspace approach for training neural networks. According to our preliminary results, our algorithm appears to be faster than first-order methods for deep network training. This was made possible by carefully taking into account the local geometrical structure of the graph of the non convex error function in a suitable subspace.
 This allowed us to use different learning rates for each network layer that are automatically adjusted at each iteration. For the future, we plan to extend our algorithm to other deep architectures.

\bibliographystyle{IEEEbib}
\bibliography{refs}

\end{document}